\documentclass[10pt,twocolumn,letterpaper]{article}

\usepackage{wacv}              % To produce the CAMERA-READY version
\usepackage{graphicx}
\usepackage{amsmath}
\usepackage{amssymb}

\usepackage{multirow} % add
\usepackage{booktabs}
\usepackage{algorithm}
\usepackage{algorithmic}

\usepackage[pagebackref,breaklinks,colorlinks]{hyperref}

\usepackage[capitalize]{cleveref}
\crefname{section}{Sec.}{Secs.}
\Crefname{section}{Section}{Sections}
\Crefname{table}{Table}{Tables}
\crefname{table}{Tab.}{Tabs.}

\def\wacvPaperID{885} % *** Enter the WACV Paper ID here
\def\confName{WACV}
\def\confYear{2024}

\begin{document}

%%%%%%%%% TITLE - PLEASE UPDATE
\title{Improving Vision-and-Language Reasoning via Spatial Relations Modeling}

\author{Cheng Yang$^{1,\ast,}$, \ Rui Xu$^{2,\ast}$, \ Ye Guo$^{3}$, \ Peixiang Huang$^{2}$, \ Yiru Chen$^{3}$, \ Wenkui Ding$^{3}$,\  \\Zhongyuan Wang$^{3}$, \  Hong Zhou$^{1,\dag}$\\
Zhejiang University$^{1}$, Peking University$^{2}$, Kuaishou Technology$^{3}$\\
{\tt\small zijingyang@zju.edu.cn, \{xurui,huangpx\}@stu.pku.edu.cn}\\ \tt\small \{guoye03,chenyiru,dingwenkui,wangzhongyuan\}@kuaishou.com, zhouh@mail.bme.zju.edu.cn\\
$^{\ast}$\textit{(equal contribution)}\ \ $^{\dag}$\textit{(corresponding author)}
% For a paper whose authors are all at the same institution,
% omit the following lines up until the closing ``}''.
% Additional authors and addresses can be added with ``\and'',
% just like the second author.
% To save space, use either the email address or home page, not both
}

\maketitle

%%%%%%%%% ABSTRACT
\begin{abstract}
   Visual commonsense reasoning (VCR) is a challenging multi-modal task, which requires high-level cognition and commonsense reasoning ability about the real world. In recent years, large-scale pre-training approaches have been developed and promoted the state-of-the-art performance of VCR. However, the existing approaches almost employ the BERT-like objectives to learn multi-modal representations. These objectives motivated from the text-domain are insufficient for the excavation on the complex scenario of visual modality. Most importantly, the spatial distribution of the visual objects is basically neglected. To address the above issue, we propose to construct the spatial relation graph based on the given visual scenario. Further, we design two pre-training tasks named object position regression (OPR) and spatial relation classification (SRC) to learn to reconstruct the spatial relation graph respectively. Quantitative analysis suggests that the proposed method can guide the representations to maintain more spatial context and facilitate the attention on the essential visual regions for reasoning. We achieve the state-of-the-art results on VCR and two other vision-and-language reasoning tasks VQA, and NLVR$^2$.
\end{abstract}

%%%%%%%%% BODY TEXT
\section{Introduction}
\label{sec:intro}

Vision-and-language reasoning is one of the most challenging tasks in multi-modal area, and the representative benchmarks incorporate Visual Commonsense Reasoning (VCR) \cite{zellers2019recognition}, Visual Question Answering (VQA) \cite{antol2015vqa} and Natural Language for Visual Reasoning (NLVR) \cite{suhr2018corpus}. Different from VQA and NLVR, VCR task requires to select the correct answer and provide corresponding explanation simultaneously given an image-question pair. Consequently, the comprehensive cognition-level scene understanding and cross-modal reasoning are essential for VCR. With the development of vision-and-language pre-training models in recent years \cite{lu2019vilbert,chen2020uniter,yu2021ernie,li2021unimo}, state-of-the-art VCR algorithms basically follow the pretrain-and-finetune manners.

\begin{figure}
	\centering
	\includegraphics[width=0.45\textwidth]{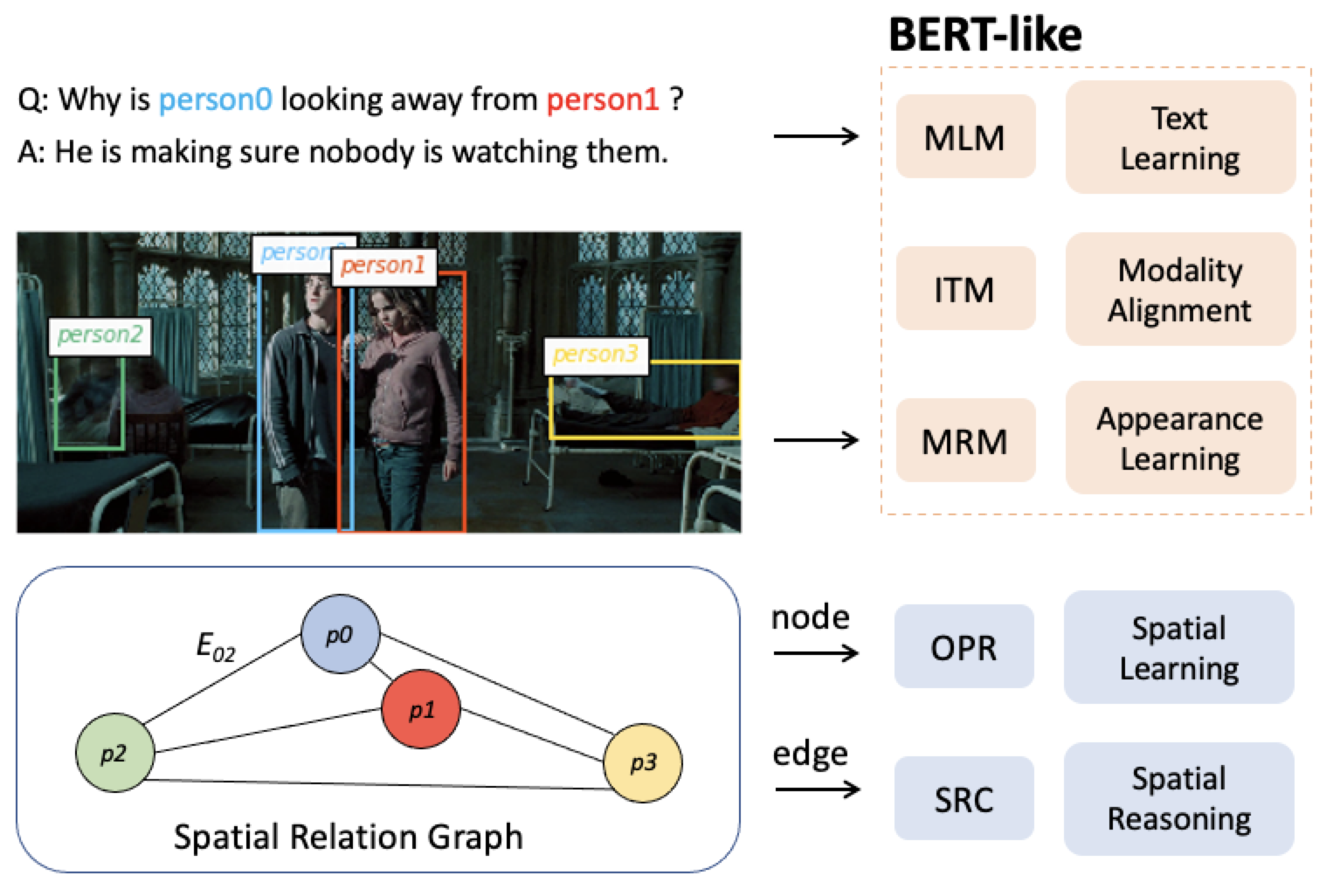}
	\caption{Existing pre-training approaches are insufficient for the excavation on visual modality. In this paper, we propose the spatial relation graph based on the visual modality and design two pre-training tasks named object position regression (OPR) and spatial relation classification (SRC) to promote the spatial context understanding and vision-and-language reasoning.}
	\label{intro}
\end{figure}

The existing vision-and-language pre-training approaches almost employ the BERT-like objectives to learn multi-modal representations , such as Masked Region Modeling (MRM) \cite{lu2019vilbert} similar to Masked Language Modeling (MLM) and Image-Text Matching (ITM) \cite{lu2019vilbert,chen2020uniter,su2019vl} similar to Next Sentence Prediction. Fig.\ref{intro} illustrates the widely applied tasks for vision-and-language pre-training. Thereinto, MLM is to predict the masked text embeddings and ITM is to distinguish the matching of the image-text pair, lacking detailed excavation on the visual modality. The task of MRM, which is designed to classify the region feature extracted by the object detector, finitely concentrates on the individual semantic category and the appearance learning. These BERT-like objectives motivated from text domain self-training are insufficient for the excavation on the complex scenario of visual modality. VCR task requires an in-depth understanding of the visual scenario and the commonsense reasoning beyond that. As Fig.\ref{intro} shows, selecting the correct answer of the visualized case basically requires the spatial-aware capture on the ``persons", which is less excavated in previous pre-training. Consequently, a more comprehensive understanding of the spatial context will significantly benefit the multi-modal reasoning for VCR.

To model the spatial context of the given scenario, we propose to construct the spatial relation graph directly with the coordinates of the visual object regions. Compared with previous methods \cite{wang2021sgeitl,li2019relation} introducing semantic-aware relations, our proposed method is free of external knowledge and training on the visual relationships. As Fig.1 shows, objects from the annotations and detection predictions constitute the graph nodes set. The value of each node is the corresponding region coordinates. The relations calculated with the coordinates of the object regions are represented as the graph edges. Beyond the BERT-like pre-training approaches, we propose to alternately learn on the constructed graph, promoting the spatial relations modeling on the multi-modal data. 

Concretely, we design two novel pre-training tasks to recover the property of nodes and edges in the constructed spatial relation graph respectively. Existing vision-and-language pre-training methods \cite{chen2020uniter,gan2020large,yu2021ernie,li2021unimo} almost adopt BUTD \cite{anderson2018bottom} to extract the object visual embeddings. Among them, the region positions are just finitely utilized as an auxiliary input to the visual embeddings. To maintain more spatial information in the multi-modal representations, we propose alternative pre-training with Object Position Regression (OPR) and Spatial Relation Classification (SRC). Taking the textual data and visual features as the context, OPR is to predict the masked positions pruning the input position information for each object. Beyond the individual spatial modeling, SRC aims to explicitly raise awareness of the spatial relations among the object. Noteworthy, the proposed alternative pre-training is significantly different with previous methods \cite{yao2018exploring,li2019relation} introducing visual relations, which followed by graph-based networks for visual representation learning. We are the first to regard the spatial relation graph as learning targets of the multi-modal pre-training, which can be easily applied on current universal and advanced transformer-based frameworks.

Experimental results demonstrate alternative training with OPR and SRC can achieve a significant performance boost compared with previous state-of-the-art methods for VCR. Quantitative analysis suggests the proposed pre-training tasks can guide the representations to capture more spatial information and improve the attention weights on more essential visual regions for reasoning. Additional experiments on VQA and NLVR$^2$ further prove the effectiveness of our method in vision-and-language reasoning field.

The contributions of our method are three-folds:
\begin{itemize}
	\item To the best of our knowledge, we are the first to regard the spatial relation graph as learning targets, and promote spatial context understanding of the vision-and-language representations.
	\item We propose two novel pre-training tasks, named Object Position Regression and Spatial Relation Classification, which can be widely applied on universal transformer-based multi-modal frameworks without external knowledge. 
	\item We achieve the state-of-the-art results 
 among the models of comparable scale. Experiments are conducted on VCR (with a significant improvement compared with previous works) and two other vision-and-language reasoning tasks VQA, and NLVR$^2$.
\end{itemize}

\begin{figure*}[h]
	\includegraphics[width=\textwidth]{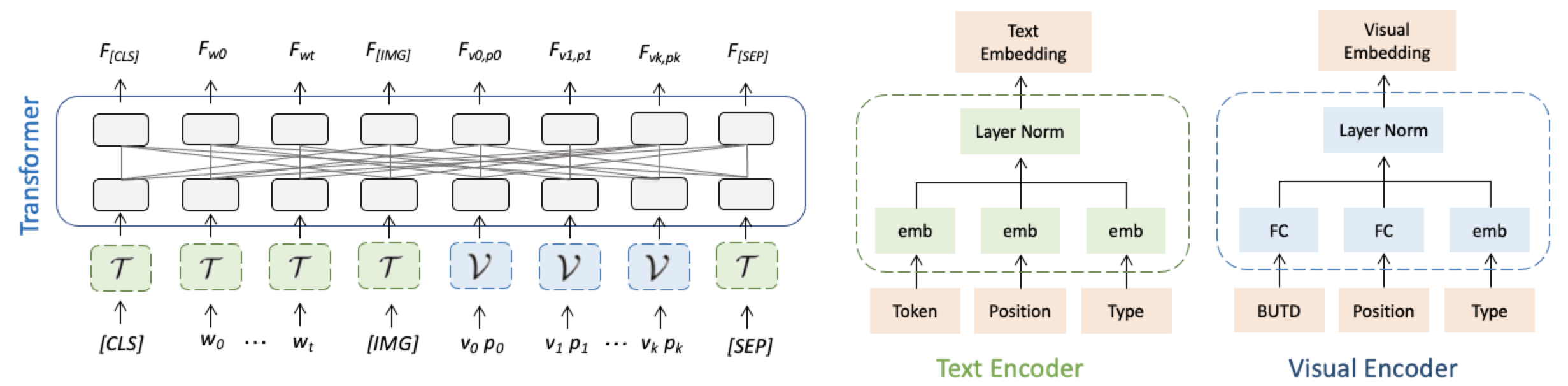}
	\caption{Overview of the model. We employ the multi-layer self-attention transformer to learn the multi-modal representation for both textual and visual data. The text encoder $\mathcal{T}$ aggregates the token, position, type of each word $w_i$ into the input text embeddings. The visual encoder $\mathcal{V}$ aggregates the object feature extracted by BUTD \cite{anderson2018bottom}, bounding box coordinates, type of each region $v_i$ into the input visual embeddings.}
	\label{overview}
\end{figure*}

\section{Related Work}

\textbf{Representation Learning.} In recent years, there are substantial interests in both vision \cite{doersch2015unsupervised,dosovitskiy2015discriminative,he2020momentum,chen2020simple} and language \cite{devlin2018bert,radford2019language,dong2019unified} pre-training for representation learning. Most visual pre-training methods are based on the convolutional neural network architecture (CNN) such as VGG \cite{simonyan2014very} and ResNet \cite{he2016deep} trained on the ImageNet dataset \cite{deng2009imagenet}. The language pre-training methods are almost based on multi-layer transformer \cite{vaswani2017attention}. BERT introduces Masked Language Modeling (MLM) pre-training task that randomly masks the input words and predicts these masked words based on the contexts. MLM has been a standard schema for linguistic model representation learning.

\noindent\textbf{Vision-and-Language Representation Learning. } ViLBERT \cite{lu2019vilbert} and LXM-ERT \cite{tan2019lxmert} are the pioneering works in vision-and-language representation learning, where two parallel transformers are utilized to process visual features or language embeddings separately, and a third transformer is built on the top for multi-modal features fusion. Compared to the above architecture, recent work such as VisualBERT \cite{li2019visualbert}, VL-BERT \cite{su2019vl}, Unicoder-V \cite{li2020unicoder} and UNITER \cite{chen2020uniter} advocate a single-stream architecture, where two modalities are fused in the early stage. VinVL \cite{zhang2021vinvl} improves the vision-and-language models by developing an improved object detection model to generate object-centric representations of images. SOHO \cite{huang2021seeing} learns to extract comprehensive image features through a visual dictionary that facilitates cross-modal understanding. CATT \cite{yang2021causal} proposes causal attention to remove the ever-elusive confounding effect in the existing attention-based models. Moreover, other techniques like knowledge integration\cite{yu2021ernie}, contrastive learning \cite{li2021unimo}, adversarial training \cite{gan2020large}, supervision from text \cite{jia2021scaling, radford2021learning}, modality matching from large scale video dataset \cite{MERLOT} are introduced to further improve the performance of the pre-trained models. The above models have brought leaping advances in vision-and-language downstream tasks such as VCR \cite{zellers2019recognition}, visual captioning \cite{zhou2020unified,qin2022pathtr,qin2023whole, huang2023assessing}, visual dialog \cite{murahari2020large} and image-text retrieval \cite{lee2018stacked}.

\noindent \textbf{Vision-and-Language Pre-training Tasks.} Most of pre-training approaches directly employ BERT-like objectives to learn multi-modal representations, such as Masked Region Model (MRM) \cite{lu2019vilbert} similar to Masked Language Model (MLM) and Image-Text Matching (ITM) \cite{lu2019vilbert,chen2020uniter,su2019vl} similar to Next Sentence Prediction (NSP). The ordinary MLM pre-training neglects the semantic relationships among the textual data. ERNIE-ViL \cite{yu2021ernie} improves the masking strategy by predicting the token types according to the textual scene graph, incorporating objects, attributes, and relationships. By increasing the corresponding masking probability, ERNIE-ViL achieves a better semantic alignment across the vision and language modality. The ordinary ITM pre-training randomly samples a negative image or text from the same training batch for each pair, leading to a coarse alignment between the textual and visual representations. To further facilitate the alignment, UNIMO \cite{li2021unimo} proposes text rewriting techniques to augment the original captions at word, phrase, and sentence levels. In this way, UNIMO obtains large volumes of positive and negative examples for each image-text pair. Furthermore, cross-modal contrastive learning (CMCL) is leveraged by UNIMO to align the textual and visual information into a unified semantic space. 

\noindent \textbf{Visual Relationship Enhanced Representation Learning.} Previous methods \cite{yao2018exploring,li2019relation,kant2020spatially,wang2021sgeitl} introduce scene graphs to model the spatial or semantic relationships among the visual objects, almost followed by graph-based attention networks to enhance the visual features. However, we propose to construct the spatial relation graph without external semantic information, are the first to regard the constructed spatial graph as learning targets of multi-modal pre-training, promoting spatial context understanding of the vision-and-language representations. Our proposed spatial relations modeling can be easily applied on large-scale and universal transformer frameworks with designed masked strategies and loss functions. Experimental results demonstrate that our method gains significant improvement on several benchmarks.

\section{Approach}

In this section, we first introduce the architecture of our model. Then we illustrate the construction of spatial relation graph and the proposed pre-training tasks for spatial relations modeling. Finally, we describe the complete pre-training objectives and procedures with alternative learning.

\subsection{Overview of the Pre-trained Model}

The vision-and-language pre-trained model aims at learning the joint representations that integrate information of both visual and textual modalities. As shown in Fig.\ref{overview}, we employ multi-layer transformer \cite{devlin2018bert} to learn the unified representations. For texts data $\mathbf{w}{=}\{w_i\}$, we adopt the token embedding initialized by RoBERTa \cite{liu2019roberta}, and special tokens incorporating $[CLS]$, $[IMG]$ and $[SEP]$ are added to the tokenized sequences. The texts are divided into different types for the questions or answers. With the text encoder $\mathcal{T}$ in Fig.\ref{overview} shows, the input text embedding for each sub-word is generated by aggregating its original token embedding, sequence position embedding, and type embedding. 

Similarly, the image is also converted to a sequence of visual embeddings. Consistent with \cite{chen2020uniter,yu2021ernie,li2021unimo}, we use BUTD \cite{anderson2018bottom} to detect the foreground regions and extract the visual features correspondingly, denoted as $\mathbf{p}{=} \{p_i\}$ and $\mathbf{v}{=}\{v_i\}$ respectively. The position information for each object is encoded via a 5-dimensional vector $p_i$ as Equation (1) shows, where $(x_1, y_1)$ and $(x_2, y_2)$ denote the region top-left and bottom-right corner coordinates, while $W$ and $H$ denote the width and height of the image. The position vectors are projected to the same dimension as $\mathbf{v}$. With the visual encoder $\mathcal{V}$, the input visual embedding for each region is generated by aggregating its visual feature, position embedding, and the visual type embedding. For a text-image pair, its textual and visual tokens are concatenated as a sequence. Then the sequence input embeddings are feed into the multi-layer transformer to learn the final multi-modal representations: $\{F_{[CLS]},F_{w0},\dots,F_{w_t},F_{[IMG]},F_{v_0, p_0},\dots,F_{v_k, p_k},F_{[SEP]}\}$. 

\begin{equation}
	p_i=(\frac{x_1}{W}, \frac{y_1}{H}, \frac{x_2}{W}, \frac{y_2}{H}, \frac{(y_2-y_1)(x_2-x_1)}{WH})
\end{equation}

\subsection{Spatial Relations Modeling}

The existing vision-and-language pre-training tasks partially concentrate on the individual category and appearance learning. For a complex visual scenario, objects positions and the interactive spatial relations contain more information to be excavated for multi-modal reasoning.

\begin{figure}
	\centering
	\includegraphics[width=0.45\textwidth]{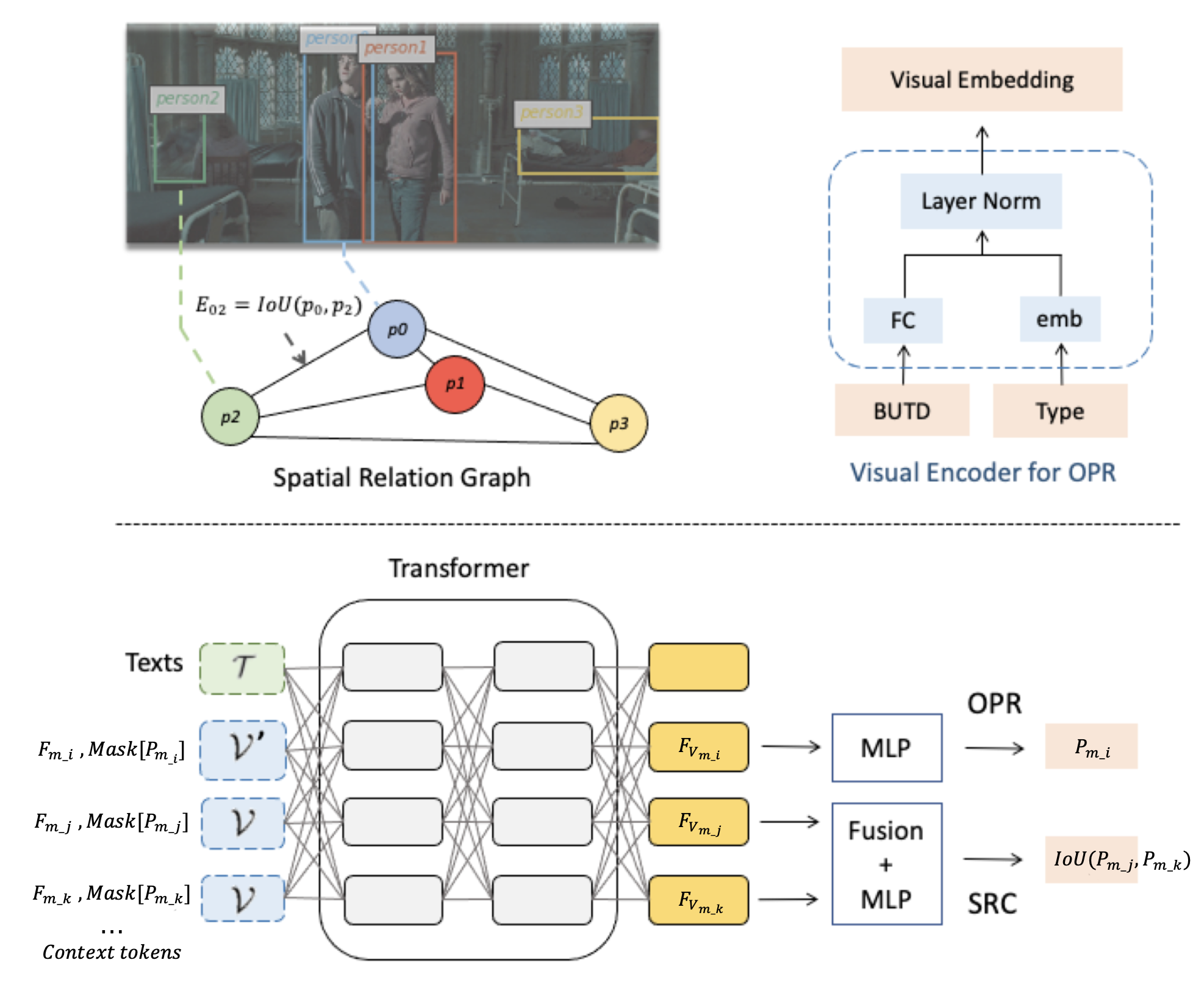}
	\caption{We propose to construct the spatial relation graph for the visual scenario. Based on the graph, we design two novel pre-training tasks named Object Position Regression (OPR) and Spatial Relation Classification (SRC) learning to recover the property of the nodes and edges respectively.}
	\label{method}
\end{figure}

\noindent{\textbf{Spatial Relation Graph.}} To model the spatial context of the scenario, we propose to construct the spatial relation graph for the given images. As Fig.3 shows, objects from annotations or detection predictions constitute the graph nodes set $\mathbf{p}$. The value of each node is the corresponding object spatial position vectors $p_i$ in Equation (1). The relations calculated with the position vectors are represented as the spatial relation edges $\{E_{ij}\}$. 

We investigate various spatial relation descriptions from the aspect of directions or overlap between the visual objects, as Table.\ref{SRC Modeling} shows. Eventually, we introduce the Intersection over Union (IoU) to quantify the spatial relations. We design two novel pre-training tasks named Object Position Regression (OPR) and Spatial Relation Classification (SRC) to learn to reconstruct the nodes and edges of the graph respectively, promoting the spatial context and comprehensive cross-model understanding.

\noindent{\textbf{Object Position Regression (OPR).}} As Fig.2 shows, in the conventional visual encoder $\mathcal{V}$, position features are finitely utilized as auxiliary information in the input embeddings. Guiding the representations to maintain more spatial information, OPR is designed to predict the position vector of each object pruning the original input with the textual and visual appearance clues as context.

Concretely, to guarantee the precision of the targeted position vector, only objects with detection confidence scores larger than 0.5 and ground truth objects are available to be masked. For OPR pre-training, we randomly mask the available position vectors with a probability 50\%. As Fig.\ref{method} shows, the visual encoder for OPR is fed with only feature embedding and token type embedding, excluding bounding box position embedding, for the position-masked objects. With additional two layers  MLP network, the transformer output $F_{v_m}$ is projected to predict the masked position vector $p_m$.  We denote the object features as $\mathbf{v}$, the corresponding position vectors as $\mathbf{p}$, and the input words as $\mathbf{w}$. Each image-text pair $(\mathbf v, \mathbf p, \mathbf w)$ is sampled from the whole training set $D$. The model parameters set is denoted as $\theta$ and the predicted position vector can be denoted as $P_{\theta}(p_m |\mathbf v,\mathbf p_{\backslash m}, \mathbf w)$. The task directly predicts the 5-dimensional vector in Equation (1). The loss function of OPR can be summarized as follows.

\begin{equation}
	\mathcal{L}_{OPR}(\theta)=\mathbb{E}_{(\mathbf v,\mathbf p, \mathbf w)\in D} ||P_{\theta}(p_m|\mathbf v,\mathbf p_{\backslash m}, \mathbf w) - p_m||^2
\end{equation}

\noindent{\textbf{Spatial Relation Classification (SRC).}} As stated above, OPR guides the pre-trained model to maintain more individual spatial information of each object, i.e nodes of the spatial relation graph in Fig.\ref{method}. For further modeling the spatial context, we propose to learn to reconstruct the edges of spatial relation graph, namely spatial relation classification (SRC).

To model the relations, we detailedly investigate the effect on the performance with different spatial metrics and modeling approaches (Table \ref{SRC Modeling}). Eventually, we introduce the IoU from the overlapping aspect to model the relationships, i.e $E_{ij}{=}IoU(p_i, p_j)$. Since IoU regression is an excessively tough task for the continuous variation, we model the prediction of the IoU as a classification problem. Concretely, we divide the IoU $\{E_{ij}\}$ to 10 classes with a uniform interval of 0.1, and the constant target is transferred to a category target $E'_{ij}$. For SRC pre-training, a pair of visual objects $(p_i, p_j)$ are sampled. The corresponding transformer output $(F_{v_i,p_i}, F_{v_j,p_j})$ are fused and projected by a two-layer MLP network to predict the relationship label $E'_{ij}$. The predicted spatial relation is denotes as $P_{\theta}(E'_{ij} |\mathbf v,\mathbf p,\mathbf w)$ and softmax-based cross-entropy loss is adopted. The loss function of SRC pre-training can be summarized as follows. 

\begin{equation}
	\mathcal{L}_{SRC}(\theta)=-\mathbb{E}_{(\mathbf v,\mathbf p, \mathbf w)\in D} log P_{\theta}(E'_{ij} | \mathbf v,\mathbf p, \mathbf w)
\end{equation}

\subsection{Alternative pre-training}

We propose to alternatively train the multi-modal model with the proposed OPR and SRC from the aspect of spatial context, combining Masked Language Modeling (MLM) and Masked Region Classification (MRC) from the aspect of semantic meanings. By the means of alternative pre-training, the model can capture more comprehensive representations for the visual and textual modality. Next, we will briefly introduce the implementation details and objectives of the adopted MLM and MRC.

For Masked Language Modeling, we randomly mask the input textual token embeddings with a probability of 15\%, and replace the masked $\mathbf w_m$ with a special token [MASK]. The model is trained to predict the masked tokens based on the surrounding context. Denote the prediction is $P_{\theta}(w_m|\mathbf v,\mathbf p, \mathbf w_{\backslash m})$, and the softmax-based cross-entropy loss function can be summarized as follows. 

\begin{equation}
	\mathcal{L}_{MLM}(\theta)=-\mathbb{E}_{(\mathbf v,\mathbf p, \mathbf w)\in D} log P_{\theta}(w_m | \mathbf v,\mathbf p, \mathbf w_{\backslash m})
\end{equation}

For Masked Region Classification, we randomly sample image regions $v_m$ and mask out their visual features with a probability of 15\%. The model is trained to predict the object category of each masked region based on the context. Additional fully-connected layers are introduced to project the prediction to the categories probability distribution $P_{\theta}(v_m|\mathbf v_{\backslash m},\mathbf p, \mathbf w)$. We adopt the KL-divergence loss function to minimize the difference between the prediction and the object detection model labeled distribution $\hat{v}_m$. The loss function of MRC can be summarized as follows.

\begin{equation}
	\mathcal{L}_{MRC}(\theta)=\mathbb{E}_{(\mathbf v,\mathbf p, \mathbf w)\in D} KL(P_{\theta}(v_m|\mathbf v_{\backslash m},\mathbf p, \mathbf w), \hat{v}_m)
\end{equation}

\section{Experiments}
\subsection{Dataset}
Visual Commonsense Reasoning (VCR) dataset \cite{zellers2019recognition} contains 100K images and 264K related questions, which are divided into the \textit{train}, \textit{val}, and \textit{test} split at a ratio of 8:1:1. The VCR task incorporates two sub-tasks: visual question answering (\textit{Q$\rightarrow$A}) and answer justification (\textit{QA$\rightarrow$R}), which are both multiple-choice problems. The holistic setting (\textit{Q$\rightarrow$AR}) requires both the chosen answer and the chosen rationale to be correct. In the visual question answering (\textit{Q$\rightarrow$A}) task, we concatenate the question and each candidate answer for the language modality. We take dot product of the final transformer outputs $F_{[CLS]}$ and $F_{[IMG]}$ to predict the matching score with an additional FC layer. For the answer justification (\textit{QA$\rightarrow$R}) task, we concatenate the question, the answer, and each candidate rationale as the input of the textual data. 

\subsection{Implementation Details}
\begin{table}[t]
	\caption{Results of the VCR task compared with the previous state-of-the-art models.}
	\renewcommand\arraystretch{1.15}
	\centering
	\resizebox{0.5\textwidth}{!}{
		\begin{tabular}{c|ccc|ccc}
			\hline
			\multirow{3}*{Models}& \multicolumn{6}{c}{VCR \textit{val}}  \\
			\cline{2-7}
			&\multicolumn{3}{c}{base}&\multicolumn{3}{c}{large} \\ 
			\cline{2-7} 
			&\;\textit{Q$\rightarrow$A}\;&\;\textit{QA$\rightarrow$R}\;&\;\textit{Q$\rightarrow$AR}\;&\;\textit{Q$\rightarrow$A}\;&\;\textit{QA$\rightarrow$R}\;&\;\textit{Q$\rightarrow$AR}\;\\
			\hline
			ViLBERT \cite{lu2019vilbert}&72.4&74.5&54.0&-&-&- \\
			VisualBERT \cite{li2019visualbert}&70.8&73.2&52.2&-&-&- \\
			SGEITL \cite{wang2021sgeitl}&-&-&-&74.9&77.2&57.8 \\
			VL-BERT \cite{su2019vl}&73.8&74.4&55.2&75.5&77.9&58.9\\
			UNITER \cite{chen2020uniter}&74.6&77.0&57.8&77.2&80.5&62.6 \\
			VILLA \cite{gan2020large}&75.5&78.8&59.8&78.5&82.6&65.2 \\
			ERNIE-ViL \cite{yu2021ernie}&76.4&79.7&61.2&79.0&83.7&66.4 \\
			Ours &\textbf{78.8}&\textbf{83.1}&\textbf{65.8}&\textbf{83.0}&\textbf{87.9}&\textbf{73.4} \\
			\hline
		\end{tabular}
	}
	\centering
	\label{main_result}
\end{table}
For fair comparison with previous methods, the experiments are conducted on two model sizes: base with 12 layers of transformer block and large with 24 layers of transformer block. We initialize the pre-trained model with UNIMO \cite{li2021unimo} for the main results, and conduct another stage of pre-training on the training split of VCR. The multi-task mix ratios for OPR, SRC, MLM and MRC are 1:1:10:1. The total number of pre-training steps is 50,000.  After the alternative pre-training, we fine-tune the model over 10,000 steps with a batch size of 64 and adopt Adamw optimizer with an initial learning rate of 6e-4.

For all experiments, we use AdamW optimizer with weight decay of $10^{-2}$. The learning rate is warmed up for 10\% of the total training steps and is decayed linearly to zero for the rest of the training. The maximum sequence length of text tokens and visual regions are set as 514 and 100, respectively. For the text tokens, we adopt Byte-Pair Encoding (BPE) to tokenize the sentence similar to RoBERTa \cite{liu2019roberta}. For the visual regions, we adopt BUTD \cite{anderson2018bottom} pre-trained on the Visual Genome \cite{krishna2016visual} to detect the object regions and extract the visual features (pooled ROI features) correspondingly, which is consistent with previous methods \cite{chen2020uniter,yu2021ernie,li2021unimo}. Specifically, regions with class detection probability exceed a confidence threshold of 0.2 are selected. For the masking strategies, we randomly mask 15\% of tokens in MLM, 15\% of object region features in MRC, and 50\% of region position vectors in OPR.

\begin{table*}[t]
	\caption{Results of VQA and NLVR$^2$ compared with previous state-of-the-art models.}
	\renewcommand\arraystretch{1.0}
	\centering
	\resizebox{\textwidth}{!}{
		\begin{tabular}{c|cc|cc|cc|cc}
			\toprule[0.75pt]
			\multirow{3}*{Models}&\multicolumn{4}{c}{VQA}&\multicolumn{4}{c}{NLVR$^2$} \\
			\cline{2-9}
			&\multicolumn{2}{c}{base}&\multicolumn{2}{c}{large}&\multicolumn{2}{c}{base}&\multicolumn{2}{c}{large} \\
			\cline{2-9}
			&\;\textit{test-dev}\;&\;\textit{test-std}\;&\;\textit{test-dev}\;&\;\textit{test-std}\;&\;\textit{dev}\;&\;\textit{testP}\;&\;\textit{dev}\;&\;\textit{testP}\;\\
			\hline
			ViLBERT \cite{lu2019vilbert}&70.6&70.9&-&-&-&-&-&- \\
			VisualBERT \cite{li2019visualbert} &70.8&71.0&-&-&67.4&67.0&-&- \\
			LXMERT \cite{tan2019lxmert}&72.4&72.5&-&-&74.9&74.5&-&- \\
			LXMERT+CATT\cite{yang2021causal}&73.5&73.7&-&-&77.2&77.2&-&- \\
			VL-BERT  \cite{su2019vl}&71.2&-&71.8&72.2&-&-&-&- \\
			UNITER  \cite{chen2020uniter} &72.7&72.9&73.8&74.0&77.2&77.9&79.1&80.0 \\
			SOHO \cite{huang2021seeing} &73.3&73.5&-&-&76.4&77.3&-&- \\
			Oscar  \cite{li2020oscar}&73.2&73.4&73.6&73.8&78.1&78.4&79.1&80.4 \\
			VILLA \cite{gan2020large} &73.6&73.7&74.7&74.9&78.4&79.3&79.8&81.5 \\
			ERNIE-ViL  \cite{yu2021ernie} &73.2&73.4&75.0&75.1&-&-&-&- \\
			UNIMO  \cite{li2021unimo}&73.8&74.0&75.1&75.3 &-&-&-&-\\
			Ours  &\textbf{74.0}&\textbf{74.2}&\textbf{75.5}&\textbf{75.6}&\textbf{79.4}&\textbf{80.1}&\textbf{80.6}&\textbf{82.2} \\
			
			\bottomrule[0.75pt]
		\end{tabular}
	}
	\centering
	\label{vqa_nlvr}
\end{table*}

\subsection{Main Results}
We compare our method against the previous state-of-the-art models and the results are illustrated in Table \ref{main_result}. We obtain a 4.6\% and 7.0\% improvement for \textit{Q$\rightarrow$AR} under the base and large setting compared with previous public state-of-the-art results achieved by ERNIE-ViL \cite{yu2021ernie}. Meantime, our result significantly outperforms SGEITL\cite{wang2021sgeitl}, which introduce external scene graphs to enhance the representation learning.

Table \ref{vcr_test} reports the \textit{test} split evaluation results of VCR. Under the large setting, we achieve the state-of-the-art results with a single model, a 2.4\% improvement even compared with the UNIMO \cite{li2021unimo} ensembling 7 models. VCR is a challenging task, which requires cross-modal commonsense reasoning and understanding on the complex scenario which is implicitly encoded in the image. Experimental results suggest we achieve new state-of-the-art results across the three benchmarks for VCR. The results strongly demonstrate the effectiveness of our method.

\begin{table}[t]
	\caption{The comparison of VCR \textit{test} set evaluation under the large setting. $*$: the result is achieved by ensembling 7 models; partially from the VCR Leaderboard \cite{zellers2019recognition}.}
	\renewcommand\arraystretch{1.15}
	\centering
	\resizebox{0.45\textwidth}{!}{
		\begin{tabular}{c|ccc}
			\hline
			\multirow{2}*{Models}& \multicolumn{3}{c}{VCR \textit{test}} \\
			\cline{2-4}
			&\;\;\textit{Q$\rightarrow$A}\;\;&\;\;\textit{QA$\rightarrow$R}\;\;&\;\;\textit{Q$\rightarrow$AR}\;\; \\
			\hline
			VL-BERT \cite{su2019vl} & 75.8 & 78.4 & 59.7 \\
			SGEITL \cite{wang2021sgeitl} &76.0&78.0&59.6 \\
			UNITER \cite{chen2020uniter} &77.3&80.8&62.8 \\
			VILLA \cite{gan2020large} &78.9&82.8&65.7\\
			ERNIE-ViL \cite{yu2021ernie} &79.2&83.5&66.3\\
			UNIMO$^*$ \cite{li2021unimo} &82.3&86.5&71.4 \\
			Ours&\textbf{83.2}&\textbf{88.1}&\textbf{73.8} \\
			\hline
	\end{tabular}}
	\centering
	\label{vcr_test}
\end{table}

\subsection{Ablation Study}
To further analyze the effectiveness of OPR and SRC, we conduct detailed ablation studies and investigate the influence on VCR at different settings.

\noindent\textbf{What type of pre-training task is more effective?} We conduct the ablation study with different pre-training tasks on UNITER-base \cite{chen2020uniter}. Previous methods \cite{chen2020uniter,yu2021ernie,gan2020large} almost adopt MLM, MRFR and MRC for the pre-training on VCR. Thereinto, MRFR is a pre-training task similar to MRC, which directly regresses the object appearance features rather than predict the semantic categories. As Table \ref{ablation} shows, the contribution of MRFR pre-training is almost useless for VCR, which suggests local appearance learning for the visual objects is already sufficient for the multi-modal pre-training.

\begin{table}[t]
	\caption{The results of VCR \textit{val} with different pre-training tasks on UNITER-base.}
	\renewcommand\arraystretch{1.15}
	\centering
        \resizebox{0.45\textwidth}{!}{
	\begin{tabular}{lccc}
		\hline
		Alternative pre-training Tasks\;\;&\textit{Q$\rightarrow$A}&\textit{QA$\rightarrow$R}&\textit{Q$\rightarrow$AR} \\
		\hline
		MLM+MRC &74.3 &76.9 &57.3\\
		MLM+MRC+MRFR& 74.3&76.9&57.4 \\
		MLM+MRC+OPR& 74.6 & 77.2&58.0 \\
		MLM+MRC+SRC& 74.8&77.1 & 58.1 \\
		MLM+MRC+SRC+OPR& \textbf{75.2}&\textbf{77.9}&\textbf{58.7} \\
		\hline
	\end{tabular}}
	\label{ablation}
\end{table}

However, the performance is significantly better if MRFR is replaced with either SRC or OPR. As Table.\ref{ablation} shows, when we conduct SRC and OPR simultaneously, a 1.0\% improvement on QA$\rightarrow$R and a 1.4\% improvement on Q$\rightarrow$AR can be obtained. Experimental results suggest spatial relations modeling is a valuable and vital complement for current multi-modal representation learning on visual modality. For further verification on the general effectiveness of OPR and SRC, we apply the proposed method on UNITER-large and VILLA-large \cite{gan2020large}. As Table \ref{ablation_2} shows, both UNITER and VILLA can obtain an improvement with OPR and SRC alternative pre-training. The ablation study on various pre-trained models convincingly demonstrates pre-training from the spatial perspective is necessary and effective for the multi-modal reasoning tasks.

\begin{table}[t]
	\caption{The results of VCR \textit{val} with OPR and SRC on UNTER and VILLA-large.}
	\renewcommand\arraystretch{1.15}
	\centering
        \resizebox{0.45\textwidth}{!}{
	\begin{tabular}{lccccccc}
		\hline
		\multicolumn{3}{l}{Models}&\textit{Q$\rightarrow$A}& &\textit{QA$\rightarrow$R}& &\textit{Q$\rightarrow$AR} \\
		\hline
		UNITER && &77.1&&80.3&&62.1 \\
		UNITER+OPR+SRC&& &\textbf{78.3}&&\textbf{81.6}&&\textbf{64.2} \\
		\hline
		VILLA && &78.1&&82.0&&64.4      \\
		VILLA+OPR+SRC& & &\textbf{78.4}&&\textbf{82.3}&&\textbf{64.8} \\
		\hline
	\end{tabular}}
	\label{ablation_2}
\end{table}

\noindent\textbf{How to model the spatial relations?} We investigate different metrics for the spatial relations and various modeling approaches for SRC, as Table \ref{SRC Modeling} shows. For direction prediction, we conduct left/right and upside/below classification between the visual object centers. For overlapping prediction, SRC learns to classify whether the sampled object regions are overlapped. Experimental results suggest overlapping prediction is a better metric than direction prediction. The supposed reason is the overlapping can reflect more relevance of the objects. 

Further, we conduct more detailed investigation on the overlapping prediction with IoU and GIoU \cite{rezatofighi2019generalized}. Experimental results suggest fine-grained classification on IoU can obtain a superior performance than binary classification on overlapping. In contrast, IoU regression is a excessively tough task, even harmful to the final performance. The performance of GIoU classification is slightly weaker than the trivial IoU metric, eventually we select IoU classification as the pre-training objective for SRC.

\begin{table}[t]
	\caption{The results of VCR \textit{val} with different SRC modeling methods.}
	\renewcommand\arraystretch{1.15}
	\setlength\tabcolsep{3pt}
	\centering
        \resizebox{0.45\textwidth}{!}{
	\begin{tabular}{lccccccccc}
		\hline
		\multicolumn{6}{l}{SRC Modeling}&\textit{Q$\rightarrow$A}&\textit{QA$\rightarrow$R}&\textit{Q$\rightarrow$AR} \\
		\hline
		Direction Prediction &&&&&&81.3&87.1&71.0& \\
		Overlapping Prediction &&&&&&81.5&87.0&71.2\\ \hline
		IoU Regression  &&&&&& 81.2&86.9&70.8\\
		IoU Classification &&&&&& \textbf{81.7}&\textbf{87.1}&\textbf{71.3} \\
		GIoU Classification &&&&&& 81.3&\textbf{87.1}&71.1 \\
		\hline
	\end{tabular}}
	\label{SRC Modeling}
\end{table}

\subsection{Experiments on Other Dataset}

To prove the generalization of our method, we conduct experiments on other vision-and-language reasoning tasks, Visual Question Answering (VQA) \cite{antol2015vqa} and Natural Language for Visual Reasoning (NLVR) \cite{suhr2018corpus}. Results compared with the state-of-the-art models are summarized in Table.\ref{vqa_nlvr}.

\noindent \textbf{Visual Question Answering (VQA).} VQA2.0 contains 204K images and 1.1M related questions, which are divided into the \textit{train}, \textit{val}, and \textit{test} split at a ratio of 2:1:2. The VQA task requires answering natural language questions according to the given images. We treat VQA as a multi-label classification task assigning a soft target score to each answer based on its relevancy to the 10 human answer responses. We take dot product of the outputs $F_{[CLS]}$ and $F_{[IMG]}$ and map the representations into 3,129 possible answers with an additional two FC layers. We adopt the same pre-training schedule as VCR. Fine-tuning on VQA is performed over 5K steps with a batch size of 256 and we adopt the Adamw optimizer with an initial learning rate of 5e-4. 

As Table.\ref{vqa_nlvr} shows, we achieve new state-of-the-art results both on base and large setting. Concretely, we get a 0.2\% performance boost under the base model setting evaluated on VQA \textit{test-dev} and \textit{test-std}. For the large model setting, the margin reaches up to 0.4\% compared with UNIMO \cite{li2021unimo}. The results suggest the spatial relations modeling is also effective for question-answer tasks.

\noindent \textbf{Natural Language for Visual Reasoning.} NLVR$^2$ contains 107K examples of human-written English sentences, which are divided into the train, \textit{dev}, \textit{testP} and \textit{testU} at a ratio of 12:1:1:1. The task is to determine whether a natural language caption is corresponding with a series of photographs. We take dot product of the final transformer outputs $F_{[CLS]}$ and $F_{[IMG]}$ to predict the matching score for each image-text pair with an additional FC layer. For fine-tuning, we train the models with 5K steps totally and a batch size of 32. The Adamw optimizer with an initial learning rate 2e-5 is adopted. 

As Table \ref{vqa_nlvr} shows, we achieve a 1.0\% and 0.8\% improvement on NLVR$^2$ \textit{dev} under the base and large setting respectively. For NLVR$^2$ \textit{testP}, the improvement is 0.8\% and 0.7\% correspondingly. Validation on NLVR$^2$ also support the conclusion stated above.

\begin{table}[t]
	\caption{Average correlation coefficient between the position embeddings and the transformer input visual / output $F_{[CLS]}$ representations.}
	\renewcommand\arraystretch{1.15}
	\centering
	\resizebox{0.45\textwidth}{!}{
		\begin{tabular}{lccccc|cc}
			\hline
			\multicolumn{5}{l}{Alternative pre-training}&\;\;Input Corr.\;\;&\;\;Output Corr.\;\;\\
			\hline
			MLM+MRC&&&&& 0.14&-0.0060 \\
			MLM+MRC+SRC&&&&& \textbf{0.19}&\ 0.0009 \\
			MLM+MRC+OPR&&&&& 0.16&\ 0.0020 \\
			MLM+MRC+OPR+SRC&&&&&0.19&\ \textbf{0.0041} \\
			\hline
	\end{tabular}}
	\label{Corr Analysis}
\end{table}

\begin{figure*}[t]
	\centering
	\includegraphics[width=0.70\textwidth]{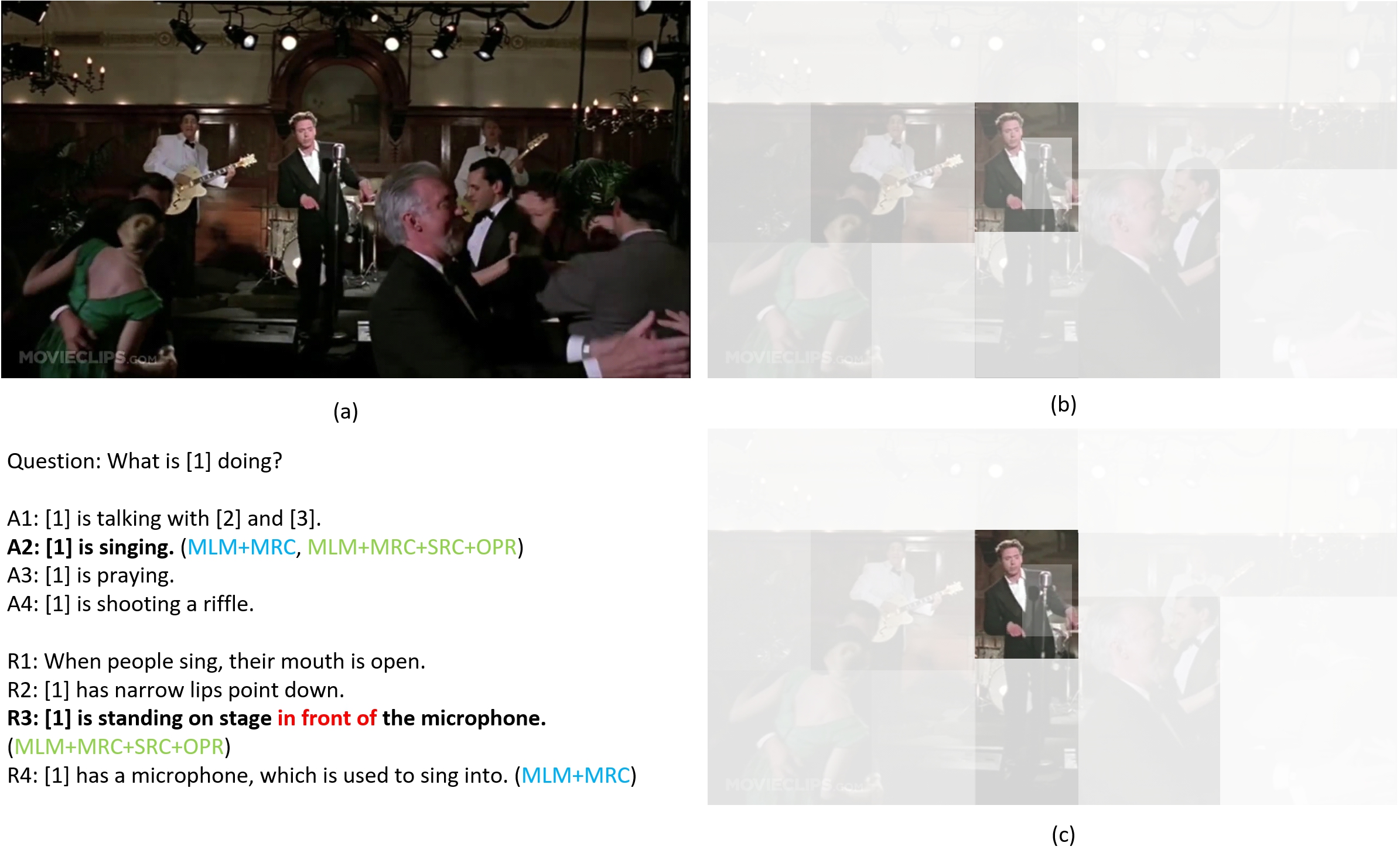}
	\caption{Case study on the text-to-object attentions (darker represents larger). (b) is the attention visualization of baseline, (c) is the attention with OPR and SRC pre-training. The correct answer and rationale are marked in bold. The answers picked by the models are indicated in parenthesis. pre-training with OPR and SRC, which memorizing more spatial relations clues in the visual scenario, can improve the attention weights on the more essential visual regions, thus benefits the multi-modal reasoning.} 
	\label{visualization}
\end{figure*}

\subsection{Discussion}

In this section, we discuss the impact of spatial-aware modeling on the multi-modal representation learning and reasoning. By quantitative analysis on the features and attention weights, the motivation can be further demonstrated.

\noindent\textbf{How spatial modeling impacts the representation learning?} To explore how the proposed spatial perspective modeling impacts on the multi-modal representations, we conduct feature correlation analysis on 100 random VCR samples as Table \ref{Corr Analysis} shows. Thereinto, the ``Input Corr" denotes the average correlation coefficient between the input position embedding and the input visual representation $\mathcal{V}(v_k, p_k)$ for each object. The ``Output Corr" denotes the average correlation coefficient between the input position embeddings and the output multi-modal representation $F_{[CLS]}$. 

Quantitative results suggest the correlation coefficient stated above is boosted with OPR and SRC alternative pre-training. It can be inferred that the proposed spatial relations modeling can facilitate the maintaining and memorization of the spatial context before and after the transformer layers by alternative pre-training. The spatial information eventually can be exploited in the fine-tuning stage, thus benefits the multi-modal reasoning downstream tasks.

\noindent\textbf{Attention weights analysis.} Fig.\ref{visualization} provides an example of the learned text-to-object attentions. We can see that the baseline model selects the right answer, but the wrong rationale, which can be corrected with the proposed OPR and SRC pre-training. In this case, the spatial relation between the person[1] and the microphone is essential for the reasoning. The attention visualization suggests that pre-training with OPR and SRC, which memorizing more spatial relations clues, can improve the attention weights on the more relevant visual regions for the multi-modal reasoning.

\section{Conclusion}

Previous vision-and-language pre-training approaches motivated by text domain are insufficient on the visual modality excavation for reasoning. To address the above issue, we propose to construct the spatial relation graph based on the given visual scenario. Further, we design two pre-training tasks named object position regression (OPR) and spatial relation classification (SRC) to learn to reconstruct the graph respectively. By alternative pre-training with OPR and SRC, we achieve state-of-the-art results on three visual-and-language reasoning downstream tasks VCR, VQA, and NLVR$^2$. In particular, even though the VCR task is considered to be a very difficult multi-modality reasoning task, our method improves the  performance of previous works by over 2.4\%, which is a significant margin. Additionally, we also conduct detailed ablative experiments to demonstrate the effectiveness of our proposed pre-training tasks. Quantitative analysis suggests the spatial relations modeling can guide the model to maintain more spatial context and facilitate the attention on essential regions, thus benefits the challenging multi-modal reasoning. \textbf{This work is supported by the National Key Research and Development Program of China (2022YFC3602601).}

%-------------------------------------------------------------------------

\newpage
%%%%%%%%% REFERENCES

{\small
\bibliographystyle{ieee_fullname}
\bibliography{egbib}
}

\end{document}